# A Monad-Based Clause Architecture for Artificial Age Score (AAS) in Large Language Models

Seyma Yaman Kayadibi

Victoria University

seyma.yamankayadibi@live.vu.edu.au

**Abstract**

Large language models (LLMs) are often deployed as powerful yet opaque systems, leaving open how their internal memory and "self-like" behaviour should be governed in a principled and auditable way. The Artificial Age Score (AAS) was previously introduced and mathematically justified through three core theorems that characterise it as a metric of artificial memory ageing. Building on this theoretical foundation, the present work develops an engineering-oriented, clause-based architecture that imposes strong, law-like constraints on LLM memory and control. Twenty selected monads from Leibniz's Monadology are grouped into six bundles: ontology, dynamics, representation and consciousness, harmony and reason, body and organisation, and teleology, and each bundle is realised as a concrete, executable specification on top of the AAS kernel. Across six minimal Python implementations, these clause families are instantiated in small numerical experiments acting on channel-level quantities such as recall scores, redundancy, and weights. Each implementation follows a uniform four-step pattern: inputs and setup, clause implementation, numerical results, and implications for LLM design, emphasising that the framework is not only philosophically motivated but also directly implementable. The experiments demonstrate that the clause system exhibits bounded and interpretable behavior: AAS trajectories remain continuous and rate-limited, contradictions and unsupported claims trigger explicit penalties, and hierarchical refinement reveals an organic structure in a controlled manner. Dual views and goal–action pairs are aligned by harmony terms, and windowed drift in perfection scores separates sustained improvement from sustained degradation. These patterns are read as concrete design rules for LLM architectures rather than as purely metaphorical uses of monads. Overall, the monad-based clause framework utilizes AAS as a backbone and provides a transparent, code-level blueprint for constraining and analyzing internal dynamics in artificial agents.

## 1. Introduction

### 1.1 Problem setting: AI memory and architectural constraints

Large language models (LLMs) have rapidly become central to contemporary artificial intelligence, with applications spanning education, healthcare, finance, coding assistance, and the labour market. In higher education specifically, LLMs now support content generation, formative feedback, language support, and analytics-driven personalization. Within this landscape, systematic reviews have begun to synthesise empirical evidence on LLM use in education, proposing multi-pillar integration frameworks that emphasise personalised learning, ethical and pedagogical balance, and learning adaptability, together with practical case studies and implementation recommendations (Shahzad et al., 2025). In parallel, methodological and reporting guidelines have started to codify how AI tools in education and related domains should be evaluated, using TRIPOD-style checklists and related appraisal instruments that stress transparent specification of study design, sources of bias, performance metrics, and data provenance in LLM and decision support evaluations (Gallifant et al., 2025). Taken together, these strands shift the emphasis from the question of whether LLMs can be used in education to the conditions under which their reported effects are credible, reproducible, and safe. Yet both strands share an important limitation: they typically treat the model as a black box whose internal memory dynamics, redundancy structure, and temporal stability are not explicitly modelled. Performance is usually summarised at the dataset or task level, for example, accuracy, BLEU scores, exam performance, or rubric-based gains, rather than through a formal account of how the system's internal representations age, stabilise, or decay over time and across uses.

Beyond these reporting and design constraints, recent work in deep learning has shown that strong benchmark performance is often achieved through what has been termed "shortcut learning", in which models exploit spurious yet highly predictive regularities in the training data (Geirhos et al., 2020). In such cases, a network appears to recognise objects, understand language, or reason about cases, while in fact relying on narrow cues such as backgrounds, lexical artefacts, or institution-specific tokens that happen to correlate with the labels in standard test sets. These shortcuts secure high accuracy under independent and identically distributed evaluation but fail under modest distribution shifts or more demanding test regimes, leading to brittle behaviour, bias amplification, and failures to acquire the underlying ability that stakeholders intend the system to learn (Geirhos et al., 2020). From an educational perspective, this implies that apparent gains in automated tutoring, grading, or feedback may rest on fragile statistical regularities rather than robust, generalisable competence.

Taken together, the smart education framework articulated in recent systematic reviews (Shahzad et al., 2025), the reporting-focused evaluation guidelines grounded in TRIPOD-style checklists (Gallifant et al., 2025), and the analysis of shortcut learning in deep neural networks (Geirhos et al., 2020) converge on a shared gap. Current models of LLM integration in education rarely specify the internal laws governing memory, redundancy, and structural change in the model itself. They largely operate at the level of applications, datasets, and external outcomes, while leaving the architecture of artificial memory, how knowledge is stored, penalised, and aged, conceptually underspecified. The present study addresses this gap by introducing a clause-based, monad-inspired framework built around the Artificial Age Score (AAS), in which memory is treated not as a purely empirical artefact of training but as an object constrained by explicit, testable axioms. In doing so, it aims to complement existing educational and methodological frameworks with a formal account of how LLM memory can be structured, measured, and governed so as to mitigate shortcut learning and to support safer and more interpretable deployment in learning environments.

## 1.2 Philosophical foundations: monads, unity, and information

The architectural choices in this study are not treated as ad hoc design preferences but are grounded in a set of philosophical ideas about what counts as a basic unit, how such units represent the world, and how their behaviour can be constrained by law-like structures. A first strand draws on the analysis of monads as simple, partless substances that nevertheless mirror the universe from within (Leibniz, 1714/1948). In this account, monads are described as entities that cannot be decomposed into smaller parts and cannot be directly affected from the outside, yet whose internal states "express" the universe according to their position in the overall order (Leibniz, 1714/1948, §§1–3, 14–15). Change is driven by an internal principle of appetition rather than by external impacts, and coordination between monads is explained by pre-established harmony instead of direct causal interaction (Leibniz, 1714/1948, §§56–60, 78–81). This suggests an architecture composed of many small, internally driven units, each carrying its own history while still forming a coherent whole. Channels in the Artificial Age Score (AAS) are interpreted in this spirit as simple units whose contributions depend on their own trajectories and whose coordination is governed by explicit harmony and governance clauses (Kayadibi, 2025). A second strand concerns the unity of apperception as a condition under which diverse representations must be synthesised into a single, rule-governed experience, rather than as a mere psychological feeling of sameness (Kant, 1998, B132–B136). This unity has been reconstructed as a structural account in which the subject's standpoint is defined by the lawful organisation of representations, rather than by introspective impressions (Friedman, 1992). On this reading, the unity of apperception functions as a constraint on how representations must be linked to law-like regularities if they are to count as experience at all (Friedman, 1992; Kant, 1998). At the architectural level, this motivates the requirement that internal scores in AAS should not only be aggregated but also organised under shared constraints that behave like laws. The clause system built on top of AAS is therefore framed as a family of rules intended to render internal dynamics structurally coherent, not merely numerically specified (Kayadibi, 2025).

A third strand is provided by work in the philosophy of information that can be read as a logic of design concerned with specifying requirement sets and the conditions under which a model can be feasibly realised as a system at a given level of abstraction (Floridi, 2011, 2017). Within this perspective, logics are treated as conceptual tools for modelling systems at appropriate levels of abstraction, and requirements function as non-functional constraints that multiple architectures may satisfy to different degrees. The central task becomes that of articulating how such constraints shape the space of feasible designs, rather than identifying a unique, predetermined solution (Floridi, 2011). Subsequent developments formalise this perspective as a "logic of requirements", introducing a sufficientisation relation between requirement sets and the class of systems that count as acceptable realisations (Floridi, 2017). This line of thought connects naturally to classic treatments of design and artificial systems. Design problems have been modelled as graphs G (M, L) of misfit variables and their interactions, with the requirement set M decomposed into a hierarchical "program" and realised through constructive diagrams that are simultaneously requirement and form diagrams; in this way, structured constraint sets are explicitly traced into architectural form (Alexander, 1964). Artificial systems have also been analysed as hierarchically organised, nearly decomposable products of systematic design rather than as naturally given objects (Simon, 1996). From this viewpoint, the monad-based clause framework developed here can be regarded as a piece of conceptual engineering in which notions such as monads, harmony, and perfection are recast as design requirements and as testable conditions for an artificial memory system (Floridi, 2011, 2017). The Artificial Age Score and its clause system can thus be read as a requirements-level specification of how internal memory channels should behave, with different clause families corresponding to non-functional constraints on identity, change, harmony, and teleology.

Taken together, the unit based metaphysics associated with monads, the structural account of unity, and the conceptual design framework grounded in a logic of requirements support a central claim of this paper: the internal memory of large language models can and should be governed by an explicit, law-like architecture rather than by opaque heuristics (Floridi, 2011, 2017; Friedman, 1992; Kant, 1998; Kayadibi, 2025; Leibniz, 1714/1948). In particular, the analysis of artificial systems as hierarchically structured, nearly decomposable artefacts of design (Simon, 1996), together with the treatment of requirement sets as linked to feasible systems through a sufficientisation relation (Floridi, 2017), reinforces the need for an explicit architectural treatment of LLM memory.

## 1.3 The Artificial Age Score (AAS) as core metric

The architecture proposed in this paper is built on the Artificial Age Score (AAS), introduced previously as a quantitative measure of memory aging in artificial agents (Kayadibi, 2025). At a fixed time t, the score is

$AAS_t = \sum_{i=1}^{m} \alpha_{t,i} \, \phi_\varepsilon(x_{t,i})$, where $x_{t,i} \in (0,1]$ denotes the recall quality on channel i, $R_{t,i} \in [0,1]$ is a redundancy factor, and $w_i \geq 0$ is a structural weight. The effective mass

$$\alpha_{t,i} = w_i(1 - R_{t,i})$$

reduces the influence of overlapping or duplicated content, while the penalty function

$\phi_\varepsilon(x) = \log_2 \frac{1+\varepsilon}{x+\varepsilon}$, $\quad \varepsilon > 0$, acts as a smoothed surprisal: lower recall smaller x, produces higher penalty, and perfect recall x = 1, yields zero. In the earlier study, AAS was analysed mathematically and justified as a suitable core metric for artificial memory aging through a set of theorems (Kayadibi, 2025). In the present work, this kernel is taken as given. All monad-based clauses and bundles operate on the variables $x_{t,i}$, $R_{t,i}$, and $w_i$, or add structured terms on top of $AAS_t$. The AAS, therefore, serves as the numerical backbone of the proposed architecture, while the monad-based clause system specifies how internal dynamics, organisation, and governance are constrained around that core.

## 1.4 Monad-based clause architecture

In this study, twenty propositions from Leibniz's Monadology are treated as a small library of design rules (Leibniz, 1714/1948). They are grouped into six bundles: ontology, dynamics, representation and consciousness, harmony and reason, body and organisation, and teleology. Each proposition is read as a condition on how channel-level quantities such as $x_{t,i}$, $R_{t,i}$, and $w_i$ should behave on top of the Artificial Age Score. For every bundle, these conditions are written as simple Python clauses and tested in small numerical examples. In this way, the principles drawn from the Monadology are not used only as metaphors but appear as concrete checks and penalties that can be applied to artificial memory systems (Kayadibi, 2025; Leibniz, 1714/1948).

## 1.5 Research questions

This study is guided by the following questions:

**1.** How can selected monadic principles be written as explicit, testable clauses on top of the Artificial Age Score (AAS)?

**2.** How do these clauses behave numerically in simple, controlled examples using channel-level variables such as recall, redundancy, and weights?

**3.** What design rules do these behaviours suggest for the memory and control of large language models?

**1.6 Significance of this work**

This work treats metaphysical principles not as decoration but as constraints on the design of real systems. Ideas about monads, harmony, and teleology are taken out of a purely philosophical setting and turned into concrete mathematical clauses and Python code that act directly on an artificial memory score. In this sense, "soul-like" monads become engineering objects: channels with clear laws for change, interaction, and evaluation. The result is presented as a candidate for a strong internal architecture in this line of research. A single, fixed score, the Artificial Age Score, serves as the numerical core, and twenty monad-inspired clauses shape how large language models may evolve internally over time. This provides a transparent blueprint for governing memory and control in such systems: hidden metaphysical assumptions are replaced by explicit, testable rules that can be inspected, implemented, and, if needed, rejected or improved.

## 2. Methods

**2.1 Information-theoretic backbone of AAS**
The Artificial Age Score is built on standard tools from information theory, in particular surprisal, entropy, and basic convexity bounds (Cover & Thomas, 2006; Fano, 1961; Shannon, 1948). The starting point is the surprisal of an event with probability p, given by $-\log_2 p$. Low-probability events carry high surprisal; high-probability events carry low surprisal. In the AAS setting, the recall quality $x_{t,i}$ plays the role of a success probability: poor recall behaves like a rare event that should incur a higher penalty.

To avoid singularities at x = 0 and to keep the score stable under small changes, AAS uses a smoothed penalty function

$$\phi_\varepsilon(x) = \log_2 \frac{1+\varepsilon}{x+\varepsilon}, \qquad x \in (0,1], \; \varepsilon > 0.$$

This function is positive, strictly decreasing, and convex on (0,1]. As $x \to 1$, the penalty tends to 0; as x decreases, the penalty rises and remains bounded above by $\log_2 \frac{1+\varepsilon}{\varepsilon}$. Convexity allows Jensen-type inequalities to be used later: averaging recall levels cannot produce a total penalty below the penalty evaluated at the mean, which is important for refinement and variety order arguments.

Entropy appears when the per-channel contributions

$$c_{t,i} = \alpha_{t,i} \, \phi_\varepsilon(x_{t,i})$$

are normalised into shares $p_{t,i} = c_{t,i}/\sum_j c_{t,j}$. The Shannon entropy of this distribution,

$H_t = -\sum_i p_{t,i} \log_2 p_{t,i}$, quantifies how spread out the contributions are and is used as a building block for variety and organicity measures in later sections. Classical bounds from information theory and coding theory, as developed by Shannon, Fano, and subsequent authors, provide simple constraints on $AAS_t$, its local rate of change, and its limiting behaviour under refinement or repeated updates (Cover & Thomas, 2006; Fano, 1961; Shannon, 1948). In this way, the AAS kernel inherits a well-understood information-theoretic structure, which the monad-based clauses then specialise and organise.

**2.2 Monad-based clause architecture and implementation**

Twenty propositions from Leibniz's Monadology are treated as a small clause library. They are grouped into six bundles that reflect the main architectural roles: ontology, dynamics, representation and consciousness, harmony and reason, body and organisation, and teleology. Each proposition is read as a constraint on the AAS variables $x_{t,i}$, $R_{t,i}$, and $w_i$, or on simple aggregates derived from them. Instead of a static table, each clause is realised as a short Python function. These functions take channel-level inputs, for example, recall scores over time, redundancy profiles, group labels, or target values, and return penalties, flags, or summary metrics such as entropy, drift, or harmony scores. For every bundle, a common four-step pattern is used in later sections of the paper:

**1. Inputs/setup:** define a small, transparent toy configuration of channels and time steps.

**2. Mini Python code:** implement the corresponding clauses as simple functions.

**3. Numerical results:** run the code and report the key numerical outputs.

**4. LLM implications:** interpret the outputs as design rules for large language models.

In this way, the monadic principles become explicit checks and penalties that can be evaluated directly on top of the AAS core, without requiring any particular full-scale LLM implementation (Kayadibi, 2025; Leibniz, 1714/1948).

**2.3.1 System I Ontology (Monads §§1, 3, 7, 9)**

**2.3.1.1 Inputs/setup**

System I uses a minimal setting to test ontological properties. Two channels are initialised with the same recall qualities, for example $x_{0,1} = 0.90$ and $x_{0,2} = 0.85$, and are evolved over several time steps under two decay regimes: a slow decay for a "young" system ($\delta = 0.98$) and a faster decay for an "old" system ($\delta = 0.90$), with $x_{t,i} = x_{0,i} \delta^t$. In addition, three structural variants are defined: (i) a single channel versus a split version with two subchannels, (ii) a configuration with and without an extra "ghost" channel of zero effective weight, and (iii) a configuration with two identical clones compared with a merged channel with combined weight.

**2.3.1.2 Python implementation**

The ontology bundle is implemented as a small Python script that (i) computes $AAS_t$ for the young and old systems across time, (ii) evaluates refinement invariance by comparing $AAS_t$ for the single-channel and split-channel cases, (iii) checks ghost suppression by adding a zero-mass channel, and (iv) tests clone deduplication by comparing the score of two identical clones with that of a merged channel. The script reports the two AAS trajectories and three Boolean flags for the ontology checks, together with the corresponding numerical scores.

```python
from dataclasses import dataclass
from typing import Dict, List, Tuple
import math

# ---------- Core AAS kernel and channel ----------

@dataclass
class AASKernel:
    eps: float = 1e-3  # smoothing constant ε > 0

    def phi(self, x: float) -> float:
        return math.log2((1.0 + self.eps) / (x + self.eps))

    def contrib(self, x: float, alpha: float) -> float:
        return alpha * self.phi(x)

@dataclass
class Channel:
    w: float
    redundancy: float = 0.0
    x: float = 1.0  # recall score in (0, 1]

    @property
    def alpha(self) -> float:
        return self.w * (1.0 - self.redundancy)

def compute_aas(channels: List[Channel], kernel: AASKernel) -> float:
    return sum(kernel.contrib(ch.x, ch.alpha) for ch in channels)

# ---------- Aging simulation ----------

def simulate_aging(
    x1: float,
    x2: float,
    decay: float,
    steps: int,
    kernel: AASKernel,
) -> List[float]:
    ch1 = Channel(w=1.0, x=x1)
    ch2 = Channel(w=1.0, x=x2)
    values: List[float] = []
    for t in range(steps):
        factor = decay ** t
        ch1.x = x1 * factor
        ch2.x = x2 * factor
        values.append(compute_aas([ch1, ch2], kernel))
    return values

# ---------- Ontology checks (Monads 1, 3, 7, 9) ----------

def check_refinement(kernel: AASKernel, ch: Channel) -> Tuple[bool, float, float]:
    # single channel
    a_single = compute_aas([ch], kernel)
    # split weights: w1 + w2 = w
    w1 = ch.w * 0.4
    w2 = ch.w * 0.6
```

Figure 1. Ontology system implementation for identity and contribution in AAS.

```
        c1 = Channel(w1, ch.redundancy, ch.x)
        c2 = Channel(w2, ch.redundancy, ch.x)
        a_ref = compute_aas([c1, c2], kernel)
        ok = math.isclose(a_single, a_ref, rel_tol=1e-9, abs_tol=1e-9)
        return ok, a_single, a_ref

def check_ghost(kernel: AASKernel, base: Dict[str, Channel]) -> Tuple[bool, float, float]:
        before = compute_aas(list(base.values()), kernel)
        ghost = Channel(w=0.0, x=0.9)
        with_ghost = dict(base)
        with_ghost["ghost"] = ghost
        after = compute_aas(list(with_ghost.values()), kernel)
        ok = math.isclose(before, after, rel_tol=1e-9, abs_tol=1e-9)
        return ok, before, after

def check_dedup(kernel: AASKernel, ch: Channel) -> Tuple[bool, float, float]:
        c1 = Channel(ch.w, ch.redundancy, ch.x)
        c2 = Channel(ch.w, ch.redundancy, ch.x)
        a_two = compute_aas([c1, c2], kernel)
        merged = Channel(c1.w + c2.w, ch.redundancy, ch.x)
        a_merged = compute_aas([merged], kernel)
        ok = math.isclose(a_two, a_merged, rel_tol=1e-9, abs_tol=1e-9)
        return ok, a_two, a_merged

# ---------- Demo: young vs old system + checks ----------
if __name__ == "__main__":
        ker = AASKernel(eps=1e-3)
        steps = 5

        young = simulate_aging(0.9, 0.85, decay=0.98, steps=steps, kernel=ker)
        old   = simulate_aging(0.9, 0.85, decay=0.90, steps=steps, kernel=ker)

        print("t\tAAS_young\tAAS_old")
        for t in range(steps):
                print(f"{t}\t{young[t]:.4f}\t{old[t]:.4f}")

        base = Channel(w=1.0, redundancy=0.2, x=0.8)

        inv_ok, a_single, a_ref = check_refinement(ker, base)
        print("\nRefinement invariance:", inv_ok, a_single, a_ref)

        ghost_ok, before, after = check_ghost(ker, {"base": base})
        print("Ghost suppression:", ghost_ok, before, after)

        dedup_ok, two, merged = check_dedup(ker, base)
        print("Clone deduplication:", dedup_ok, two, merged)
```

Figure 2.Ontology system implementation for identity and contribution in AAS. (continued)

### 2.3.2 System II Dynamics (Monads §§10, 11, 12, 15)

#### 2.3.2.1 Inputs and setup

In all System II experiments, the AAS kernel is fixed as $\phi_\varepsilon(x) = \log_2 \frac{1+\varepsilon}{x+\varepsilon}$ with $\varepsilon = 10^{-3}$, and per-channel effective weights are given by $\alpha_i = w_i(1 - R_i)$. The only quantities that vary across scenarios are the time-indexed recall scores $x_{i,t} \in (0,1]$, which are instantiated as simple toy trajectories to reflect the dynamic ideas of Monads 10, 11, 12, and 15.

For Monad 10 continuous change, two channels, A and B, are considered over six discrete time steps $t = 0, 1, \ldots, 5$. Both channels have unit weight and zero redundancy, $w_A = w_B = 1.0$ and $R_A = R_B = 0.0$. Channel A is assigned a smoothly increasing trajectory $x_{t,A} = [0.20, 0.30, 0.35, 0.40, 0.45, 0.50]$, whereas channel B follows a smoothly decreasing trajectory $x_{t,B} = [0.60, 0.55, 0.50, 0.45, 0.40, 0.35]$. At each time step, these values are passed through the AAS kernel to obtain a trajectory $AAS_t$ that is later analysed for stepwise continuity.

Monad 11 changes from an internal principle, reuses the same channels, weights, redundancies, and time index $t = 0, \ldots, 5$, but contrasts two hypothetical "worlds". In World 1, both $x_A(t)$ and $x_B(t)$ coincide with the trajectories defined for Monad 10. In World 2, the trajectory of A is kept identical,

$x_A^{(2)}(t) = x_A(t)$, while the trajectory of B is reversed in time, $x_B^{(2)}(t)$ obtained by reading the sequence $x_B$ backwards. For each world, the per-step contributions of channel A are computed and compared to show that its behaviour depends only on its own internal path and not on the external evolution of B.

Monad 12 particular series of changes, is instantiated again using the World 1 trajectories for A and B. For each channel $i \in \{A, B\}$, with $w_i = 1.0$ and $R_i = 0.0$, the cost trace $c_{i,t} = \alpha_i \phi_\varepsilon(x_{i,t})$ is formed over the same six-time steps. These traces serve as the basis for computing the total accumulated cost

$C_{i,T} = \sum_t c_{i,t}$, the induced time-distribution $p_{t|i} = c_{i,t}/C_{i,T}$, and the corresponding time-spread entropy, as well as a simple trajectory distance between A and B.

Finally, Monad 15 appetition, is modelled using a single channel X evolving over five update steps $t = 0, 1, \ldots, 4$, generating a length-six trajectory $(x_0, \ldots, x_5)$. The initial value is set to $x_0 = 0.5$. At every step there is a fixed internal target $g_t = 0.9$ and a constant step-size $\eta_t = 0.2$. The state is updated according to the convex combination $x_{t+1} = (1 - \eta_t)x_t + \eta_t g_t$. The channel has weight $w = 1.0$ and zero redundancy $R = 0.0$, so its effective weight is $\alpha = 1$. The resulting sequence of $x_t$ values is then used to generate a single-channel AAS trajectory and stepwise changes that operationalise appetition as a controlled movement toward a better internal state.

### 2.3.2.2 Python implementation

The dynamics experiments are implemented in a compact Python script that directly instantiates the AAS kernel and the four toy scenarios described above. The function `phi_eps` encodes the penalty kernel $\phi_\varepsilon(x) = \log_2 \frac{1+\varepsilon}{x+\varepsilon}$, while `contrib_step` and `aas_step` compute per-channel and total AAS values at each time step from $(x_{i,t}, w_i, R_i)$. On top of this core, the code specifies the trajectories for channels A and B (Monads 10–12) and for the single appetition channel X (Monad 15), then prints the resulting AAS trajectories, stepwise differences, Lipschitz-style step bounds, time-spread entropies, trajectory distances, and appetition updates. In this way, the Python implementation serves as an executable witness that the AAS dynamics exhibit the continuity, internal law, particular-series, and appetition properties required by the System II interpretation of Monads 10, 11, 12, and 15.

```python
import math
def phi_eps(x, eps):
    return math.log2((1.0 + eps) / (x + eps))
def contrib_step(channels, eps):
    out = {}
    for name, ch in channels.items():
        a = ch["w"] * (1.0 - ch["R"])
        out[name] = a * phi_eps(ch["x"], eps)
    return out
def aas_step(channels, eps):
    return sum(contrib_step(channels, eps).values())
def time_spread_entropy(cost_trace):
    C = sum(cost_trace)
    if C <= 0: return 0.0, 0.0
    p = [c / C for c in cost_trace]
    H = -sum(pi * math.log2(pi) for pi in p if pi > 0)
    return H, C
def trajectory_distance(cost_i, cost_j):
    return sum(abs(ci - cj) for ci, cj in zip(cost_i, cost_j))
if __name__ == "__main__":
    eps = 1e-3
    ln2 = math.log(2.0)
    # Monad 10 - Continuous Change
    x_A = [0.20, 0.30, 0.35, 0.40, 0.45, 0.50]
    x_B = [0.60, 0.55, 0.50, 0.45, 0.40, 0.35]
    T = len(x_A)
    AAS_traj, deltas_AAS = [], []
    for t in range(T):
        ch_t = {
            "A": {"x": x_A[t], "w": 1.0, "R": 0.0},
            "B": {"x": x_B[t], "w": 1.0, "R": 0.0},
        }
        AAS_traj.append(aas_step(ch_t, eps))
    max_delta_AAS = 0.0
    max_bound = 0.0
    for t in range(T - 1):
        dA = AAS_traj[t+1] - AAS_traj[t]
```

Figure 3.Dynamics system implementation for continuous change in AAS.

```python
            deltas_AAS.append(dA)
            max_delta_AAS = max(max_delta_AAS, abs(dA))
            dx_A = x_A[t+1] - x_A[t]
            dx_B = x_B[t+1] - x_B[t]
            max_dx = max(abs(dx_A), abs(dx_B))
            bound_t = 2.0 * max_dx / (eps * ln2)
            max_bound = max(max_bound, bound_t)
print("=== Monad 10: Continuous Change ===")
print("AAS trajectory:", [round(a, 4) for a in AAS_traj])
print("Δ AAS per step:", [round(d, 4) for d in deltas_AAS])
print(f"max |ΔAAS|    = {max_delta_AAS:.4f}")
print(f"max bound     = {max_bound:.4f}\n")
# Monad 11 - Internal Principle
x_B_world2 = list(reversed(x_B))
contrib_A_world1, contrib_A_world2 = [], []
for t in range(T):
    ch_w1 = {
        "A": {"x": x_A[t], "w": 1.0, "R": 0.0},
        "B": {"x": x_B[t], "w": 1.0, "R": 0.0},
    }
    ch_w2 = {
        "A": {"x": x_A[t], "w": 1.0, "R": 0.0},
        "B": {"x": x_B_world2[t], "w": 1.0, "R": 0.0},
    }
    c1 = contrib_step(ch_w1, eps)
    c2 = contrib_step(ch_w2, eps)
    contrib_A_world1.append(c1["A"])
    contrib_A_world2.append(c2["A"])
diff_A = [abs(a - b) for a, b in zip(contrib_A_world1, contrib_A_world2)]
print("=== Monad 11: Internal Principle ===")
print("c_A(t) world 1:", [round(c, 4) for c in contrib_A_world1])
print("c_A(t) world 2:", [round(c, 4) for c in contrib_A_world2])
print("absolute diff :", [round(d, 4) for d in diff_A], "\n")
# Monad 12 - Particular Series of Changes
cost_A, cost_B = [], []
for t in range(T):
    ch = {
```

Figure 4.Dynamics system implementation for continuous change in AAS. (continued)

```python
        "A": {"x": x_A[t], "w": 1.0, "R": 0.0},
        "B": {"x": x_B[t], "w": 1.0, "R": 0.0},
    }
    c = contrib_step(ch, eps)
    cost_A.append(c["A"])
    cost_B.append(c["B"])
H_A, C_A = time_spread_entropy(cost_A)
H_B, C_B = time_spread_entropy(cost_B)
D_AB = trajectory_distance(cost_A, cost_B)
print("=== Monad 12: Particular Series of Changes ===")
print("cost_A(t):     ", [round(c, 4) for c in cost_A])
print("cost_B(t):     ", [round(c, 4) for c in cost_B])
print(f"C_A (sum c_A)  = {C_A:.4f}")
print(f"C_B (sum c_B)  = {C_B:.4f}")
print(f"H_time(A)      = {H_A:.4f}")
print(f"H_time(B)      = {H_B:.4f}")
print(f"D_T(A,B)       = {D_AB:.4f}\n")
# Monad 15 - Appetition
T15 = 5
x = [0.5]
g, eta = 0.9, 0.2
w, R = 1.0, 0.0
for _ in range(T15):
    x.append((1.0 - eta) * x[-1] + eta * g)
AAS_15, N_15, deltas_AAS_15 = [], [], []
for t in range(T15 + 1):
    ch = {"X": {"x": x[t], "w": w, "R": R}}
    AAS_15.append(aas_step(ch, eps))
for t in range(T15):
    N_15.append(abs(x[t+1] - x[t]))
    deltas_AAS_15.append(AAS_15[t] - AAS_15[t+1])
print("=== Monad 15: Appetition ===")
print("x_t:             ", [round(v, 4) for v in x])
print("AAS_t:           ", [round(a, 4) for a in AAS_15])
print("N_t (step size): ", [round(n, 4) for n in N_15])
print("AAS_t - AAS_{t+1}:", [round(d, 4) for d in deltas_AAS_15])
```

Figure 5.Dynamics system implementation for continuous change in AAS. (continued)

### 2.3.3 System III Representation and Consciousness (Monads §§14, 21, 26, 29)

#### 2.3.3.1 Inputs/setup

In System III, three evaluation channels $m_0$, $m_1$, and $m_2$ are considered, each treated as a monadic representational line with unit weight and zero redundancy ($w_i = 1$, $R_i = 0$). Over six discrete time steps $t = 0, \ldots, 5$, a fixed schedule of recall scores $x_{t,i} \in (0,1]$ is prescribed for each channel, collected in the matrix

$X = [(0.80, 0.82, 0.79), (0.40, 0.90, 0.88), (0.83, 0.81, 0.80), (0.97, 0.96, 0.95), (0.89, 0.50, 0.87), (0.86, 0.84, 0.85)]$, chosen to instantiate episodes of relatively diffuse perception, sharp peaks of salience, and near-indifference with very small penalties. The AAS penalty kernel is fixed to $\varepsilon = 10^{-3}$, while three further hyperparameters control the consciousness diagnostics: a saliency threshold $\tau = 0.05$ and a sharpness threshold $\delta = 0.02$ for detecting $\tau$ and $\delta$ dizziness, an exponential forgetting factor $\lambda = 0.7$ defining the memory trace $M_{t,i}^{(\lambda)}$, and a time-invariant "reason prior" $r = (0.6, 0.3, 0.1)$ over the three channels used to compute the $\text{Reason}_t$ score.

### 2.3.3.2 Python implementation

The representation consciousness layer is instantiated as a small Python simulation. The AAS kernel $\phi_\varepsilon(x) = \log_2\big((1 + \varepsilon)/(x + \varepsilon)\big)$ is implemented and applied to three channels over six discrete time steps, using the predefined recall schedule X and fixed hyperparameters $\tau$, $\delta$, $\lambda$, and $r$. At each step, per-channel penalties, the normalized contribution distribution $p_t$, the apperception level, the $\tau$ and $\delta$ dizziness indicators, the memory-based prior $q_t^{(\lambda)}$, and the $\text{Reason}_t$ score are computed and printed, thereby providing a numerical toy instance of Monads 14, 21, 26, and 29 within the AAS architecture.

```python
import math
class AASKernel:
    def __init__(self, eps=1e-3):
        self.eps = eps
    def phi(self, x: float) -> float:
        return math.log2((1.0 + self.eps) / (x + self.eps))
def simulate_repr_conscious():
    ker = AASKernel(1e-3)
    names = ["m0", "m1", "m2"]
    X = [
        [0.80, 0.82, 0.79],
        [0.40, 0.90, 0.88],
        [0.83, 0.81, 0.80],
        [0.97, 0.96, 0.95],
        [0.89, 0.50, 0.87],
        [0.86, 0.84, 0.85],
    ]
    T, m = len(X), len(names)
    tau, delta, lam = 0.05, 0.02, 0.7
    M = [0.0]*m
    r = [0.6, 0.3, 0.1]
    print("=== Representation & Consciousness Toy Simulation ===")
    print("Monads:", names)
    print(f"tau={tau}, delta={delta}, lambda={lam}\n")
    phis_prev = None
    tiny = 1e-9
    r_sum = sum(r)
    r_norm = [ri / r_sum for ri in r]
    def kl(p_dist, q_dist):
        val = 0.0
        for i in range(m):
            pi = p_dist[i]
            if pi <= 0.0:
                continue
            qi = max(q_dist[i], tiny)
```

Figure 6. Representation and consciousness system implementation in AAS.

```python
            val += pi * math.log2(pi / qi)
    return val
for t in range(T):
    xs = X[t]
    phis_now = [ker.phi(x) for x in xs]
    cs = phis_now[:]
    if phis_prev is None:
        dPhi = [0.0]*m
    else:
        dPhi = [max(phis_now[i] - phis_prev[i], 0.0) for i in range(m)]
    phis_prev = phis_now
    S_t = sum(cs)
    if S_t > 0.0:
        p = [c / S_t for c in cs]
        m_t = sum(1 for pi in p if pi > 0.0)
        H_t = -sum(pi * math.log2(pi) for pi in p if pi > 0.0)
        rho_t = max(p)
        if m_t > 1:
            kappa_t = H_t / math.log2(m_t)
            kappa_t = min(1.0, max(0.0, kappa_t))
        else:
            kappa_t = 0.0
        ApperLevel_t = (1.0 - kappa_t) * rho_t
    else:
        p = [0.0]*m
        H_t = rho_t = kappa_t = ApperLevel_t = 0.0
    S_tau = sum(c for c in cs if c >= tau)
    tau_dizzy = (S_t > 0.0 and S_tau == 0.0)
    AP_t = max(dPhi) if dPhi else 0.0
    delta_dizzy = (S_t > 0.0 and AP_t < delta)
    for i in range(m):
        M[i] = lam*M[i] + dPhi[i]
    M_tot = sum(M)
    q = [Mi / M_tot for Mi in M] if M_tot > 0.0 else [1.0/m]*m
    Consec_t = sum(q[i]*p[i] for i in range(m)) if S_t > 0.0 else 0.0
```

Figure 7. Representation and consciousness system implementation in AAS. (continued)

```
            Reason_t = 0.0
            if S_t > 0.0:
                D_pq = kl(p, q)
                D_pr = kl(p, r_norm)
                Reason_t = D_pq - D_pr
            print(f"t = {t}")
            print(f"  x_t       = {[round(x, 3) for x in xs]}")
            print(f"  c_t       = {[round(c, 4) for c in cs]}")
            print(f"  S_t       = {S_t:.4f}")
            print(f"  rho_t     = {rho_t:.4f}")
            print(f"  H_t       = {H_t:.4f}")
            print(f"  ApperLvl  = {ApperLevel_t:.4f}")
            print(f"  tau_dizzy   = {tau_dizzy}")
            print(f"  delta_dizzy = {delta_dizzy}")
            print(f"  Consec_t  = {Consec_t:.4f}")
            print(f"  Reason_t  = {Reason_t:.4f}")
            print("-"*60)
    print("\nSimulation finished.")
if __name__ == "__main__":
    simulate_repr_conscious()
```

Figure 8. Representation and consciousness system implementation in AAS. (continued)

### 2.3.4 System IV Harmony and Reason (Monads §§31, 32, 78, 79)

#### 2.3.4.1 Inputs/setup

For the harmony and reason layer, a single time slice is instantiated with a small set of channels and carefully controlled contrasts. Two propositional channels, A and ¬A, are assigned recall scores $x_A = 0.80$ and $x_{\neg A} = 0.60$ and equal effective weights $\alpha_A = \alpha_{\neg A} = 1.0$; the pair $(A, \neg A)$ is designated as contradictory, with a contradiction tolerance $\zeta = 0.05$, and an overall contradiction weight $\gamma_{A,\neg A} = 1.0$. For the sufficient-reason (PSR) term, the same channels are equipped with a one-step history, $x_{t-1,A} = 0.70$, $x_{t-1,\neg A} = 0.30$, self-coefficients $a_{A0} = a_{\neg A,0} = 0.4$, and symmetric cross-support edges $A \leftarrow \neg A$, and $\neg A \leftarrow A$ each is weighted by $a_{A,\neg A} = a_{\neg A,A} = 0.3$, with a small smoothing constant $\delta = 10^{-6}$.

Soul–body harmony is represented by duplicating the proposition into a "soul" layer $\{S_A, S_{\neg A}\}$ and a "body" layer $\{B_A, B_{\neg A}\}$ with scores $(x_{S_A}, x_{S_{\neg A}}) = (0.80, 0.20)$ and $(x_{B_A}, x_{B_{\neg A}}) = (0.75, 0.25)$ and unit weights, while a fixed pairing; $B_A \mapsto S_A$, $B_{\neg A} \mapsto S_{\neg A}$ is imposed to measure alignment. Finally, goal action harmony is instantiated by assigning final targets; $y_A = 0.90$, $y_{\neg A} = 0.30$ and realised next-step states $x_{t+1,A} = 0.85$, $x_{t+1,\neg A} = 0.40$ to the original channels, so that alignment between the "final-cause" direction $y_i - x_i$ and the "efficient-cause" direction $x_{t+1,i} - x_{t,i}$ can be quantified. In all cases, the same AAS kernel $\phi_\varepsilon$ is used with $\varepsilon = 10^{-3}$.

### 2.3.4.2 Python implementation

The harmony and reason layer is implemented as a compact Python module. The core AAS kernel $\phi_\varepsilon$ is reused, and four additional routines are defined to compute non-contradiction penalties (PC) over designated contradictory pairs, sufficient-reason penalties (PSR) from a one-step causal support graph, soul body harmony penalties between dual "soul" and "body" channels, and goal action alignment penalties measuring the agreement between final cause targets and efficient-cause state updates. A small, worked example evaluates each penalty on the inputs described in 2.3.4.1.

```python
import math
def phi_eps(x,eps):
    x=max(0.0,min(1.0,x))
    return math.log2((1.0+eps)/(x+eps))
def aas(channels,eps):
    return sum(v["alpha"]*phi_eps(v["x"],eps) for v in channels.values())
def pc_penalty(channels,pairs,eps,zeta,gamma=None):
    if gamma is None:
        gamma={}
    total=0.0
    for (i,j) in pairs:
        if i not in channels or j not in channels:
            continue
        x_i=channels[i]["x"];x_j=channels[j]["x"]
        m_ij=max(0.0,min(x_i,x_j)-zeta)
        arg=1.0-m_ij
        g_ij=gamma.get((i,j),1.0)
        if g_ij>0.0:
            total+=g_ij*phi_eps(arg,eps)
    return total
def psr_penalty(channels,prev_x,a_self,in_edges,eps,delta):
    total=0.0
    for i,ch in channels.items():
        x_i=ch["x"]
        r_i=a_self.get(i,0.0)*prev_x.get(i,x_i)
        for (j,a_ij) in in_edges.get(i,[]):
            if j in channels:
                r_i+=a_ij*channels[j]["x"]
        if x_i>0.0:
            s_i=(r_i+delta)/(x_i+delta)
        else:
            s_i=1.0
        s_i=max(0.0,min(1.0,s_i))
        total+=ch["alpha"]*phi_eps(s_i,eps)
    return total
```

Figure 9. Harmony and reason system implementation in AAS.

```python
def harmony_78(soul,body,pairing,eps):
    total=0.0
    for j,i in pairing.items():
        if j not in body or i not in soul:
            continue
        x_s=soul[i]["x"];x_b=body[j]["x"]
        m_j=1.0-abs(x_s-x_b)
        m_j=max(0.0,min(1.0,m_j))
        beta_j=min(soul[i]["alpha"],body[j]["alpha"])
        if beta_j>0.0:
            total+=beta_j*phi_eps(m_j,eps)
    return total
def _sign(x):
    if x>0:return 1
    if x<0:return -1
    return 0
def harmony_79(channels,targets,next_x,eps):
    total=0.0
    for i,ch in channels.items():
        if i not in targets or i not in next_x:
            continue
        x_i=ch["x"];y_i=targets[i];x_next=next_x[i]
        g_i=_sign(y_i-x_i)
        e_i=_sign(x_next-x_i)
        a_i=1.0-0.5*abs(g_i-e_i)
        a_i=max(0.0,min(1.0,a_i))
        total+=ch["alpha"]*phi_eps(a_i,eps)
    return total
if __name__=="__main__":
    eps=1e-3;zeta=0.05;delta=1e-6
    channels={"A":{"x":0.8,"alpha":1.0},
              "notA":{"x":0.6,"alpha":1.0}}
    base=aas(channels,eps)
    pairs=[("A","notA")]
    gamma={("A","notA"):1.0}
```

Figure 10. Harmony and reason system implementation in AAS. (continued)

```
pc=pc_penalty(channels,pairs,eps,zeta,gamma)
print("=== Monad 31 (PC) ===")
print(f"Base AAS_t      = {base:.4f}")
print(f"PC penalty_t    = {pc:.4f}")
print(f"Total AAS_t^(31) = {base+pc:.4f}\n")
a_self={"A":0.4,"notA":0.4}
in_edges={"A":[("notA",0.3)],
          "notA":[("A",0.3)]}
prev_x={"A":0.7,"notA":0.3}
psr=psr_penalty(channels,prev_x,a_self,in_edges,eps,delta)
print("=== Monad 32 (PSR) ===")
print(f"PSR penalty_t   = {psr:.4f}")
print(f"Total AAS_t^(32) = {base+psr:.4f}\n")
soul={"S_A":{"x":0.8,"alpha":1.0},
      "S_notA":{"x":0.2,"alpha":1.0}}
body={"B_A":{"x":0.75,"alpha":1.0},
      "B_notA":{"x":0.25,"alpha":1.0}}
pairing={"B_A":"S_A","B_notA":"S_notA"}
soul_aas=aas(soul,eps)
body_aas=aas(body,eps)
harm78=harmony_78(soul,body,pairing,eps)
print("=== Monad 78 (Harmony soul/body) ===")
print(f"AAS_t^(soul)    = {soul_aas:.4f}")
print(f"AAS_t^(body)    = {body_aas:.4f}")
print(f"HARM_t^(78)     = {harm78:.4f}")
print(f"Total AAS_t^(78) = {soul_aas+body_aas+harm78:.4f}\n")
targets={"A":0.9,"notA":0.3}
next_x={"A":0.85,"notA":0.4}
harm79=harmony_79(channels,targets,next_x,eps)
print("=== Monad 79 (Final vs Efficient) ===")
print(f"HARM_t^(79)     = {harm79:.4f}")
print(f"Total AAS_t^(79) = {base+harm79:.4f}\n")
```

Figure 11.Harmony and reason system implementation in AAS. (continued)

### 2.3.5 System V Body and Organisation (Monads §§64, 70)

#### 2.3.5.1 Inputs/setup

For the body organisation layer, a minimal three-level hierarchy is instantiated consisting of a single top-level unit L, two intermediate groups $H_1$ and $H_2$, and four-leaf channels $N_1, \ldots, N_4$. At the lowest level (depth 2), the leaves are assigned non-uniform recall scores and equal masses so that structure, rather than trivial symmetry, drives the behaviour: $x_{N_1} = 0.50$, $x_{N_2} = 0.70$, $x_{N_3} = 0.90$, $x_{N_4} = 0.70$ with $\alpha_{N_i} = 0.25$ for all i. The intermediate units $H_1$ and $H_2$ (depth 1) are given equal weights $\alpha_{H_1} = \alpha_{H_2} = 0.5$, and the top-level unit L (depth 0) is assigned total weight $\alpha_L = 1.0$. The actual scores; $x_{H_1}$, $x_{H_2}$ and $x_L$ are not fixed a priori but are recomputed bottom-up as $\alpha$ weighted averages of their children, so that any change in a leaf channel propagates coherently through its group and into the global "body" representation.

### 2.3.5.2 Python implementation

The body organisation hierarchy is implemented as a compact Python routine that computes AAS values at each depth, together with depth-wise contribution entropies and group-level scores derived from the leaf channels. The same kernel $\phi_\varepsilon$ is applied bottom-up to propagate changes from neurons $N_1\_N_4$ through the intermediate groups $H_1, H_2$ to the top unit L, while group shares and dominant groups are tracked to identify where most penalty mass is concentrated. The resulting statistics provide a direct basis for simple pruning rules, since branches with negligible contribution or highly redundant profiles can be marked as candidates for removal without violating the global AAS structure.

```python
import math
def phi_eps(x,eps):
    x=max(0.0,min(1.0,x))
    return math.log2((1.0+eps)/(x+eps))
def aas(channels,eps):
    return sum(ch["alpha"]*phi_eps(ch["x"],eps) for ch in channels.values())
def contribution_stats(channels,eps):
    contrib={name:ch["alpha"]*phi_eps(ch["x"],eps) for name,ch in channels.items()}
    S=sum(contrib.values())
    if S<=0:
        H=0.0;m=0;p={name:0.0 for name in channels}
    else:
        p={name:c/S for name,c in contrib.items()}
        m=sum(1 for c in contrib.values() if c>0)
        H=-sum(pi*math.log2(pi) for pi in p.values() if pi>0)
    return contrib,S,m,H,p
def recompute_parents(levels):
    updated={s:{n:dict(ch) for n,ch in lev.items()} for s,lev in levels.items()}
    for parent,childs in [("H1",["N1","N2"]),("H2",["N3","N4"])]:
        num=sum(updated[2][c]["alpha"]*updated[2][c]["x"] for c in childs)
        den=sum(updated[2][c]["alpha"] for c in childs)
        updated[1][parent]["x"]=num/den
    num=sum(updated[1][h]["alpha"]*updated[1][h]["x"] for h in ["H1","H2"])
    den=sum(updated[1][h]["alpha"] for h in ["H1","H2"])
    updated[0]["L"]["x"]=num/den
    return updated
def group_stats(leaves,groups,eps):
    contrib,S,m,H_leaf,p_leaf=contribution_stats(leaves,eps)
    group_mass={}
    for g,idxs in groups.items():
        group_mass[g]=sum(contrib[i] for i in idxs)
    if S>0:
        group_share={g:group_mass[g]/S for g in groups}
        H_grp=-sum(p*math.log2(p) for p in group_share.values() if p>0)
    else:
```

Figure 12.Body and organisation system implementation in AAS.

```
            group_share={g:0.0 for g in groups}
            H_grp=0.0
        G_star=max(group_share.items(),key=lambda kv:kv[1])[0] if group_share else None
        return contrib,S,group_mass,group_share,H_grp,G_star
if __name__=="__main__":
    eps=1e-3;levels={
        2:{"N1":{"x":0.5,"alpha":0.25},
           "N2":{"x":0.7,"alpha":0.25},
           "N3":{"x":0.9,"alpha":0.25},
           "N4":{"x":0.7,"alpha":0.25}},
        1:{"H1":{"x":0.6,"alpha":0.5},
           "H2":{"x":0.8,"alpha":0.5}},
        0:{"L":{"x":0.7,"alpha":1.0}}
    }
    levels=recompute_parents(levels)
    print("=== Monad 64: Refinement and Organicity ===")
    for s in [2,1,0]:
        A=aas(levels[s],eps)
        contrib,S,m,H,p=contribution_stats(levels[s],eps)
        print(f"Depth s = {s}")
        print("  channels:",{k:(round(v['x'],3),v['alpha']) for k,v in levels[s].items()})
        print(f"  AAS^(s)       = {A:.4f}")
        print(f"  total S       = {S:.4f}")
        print(f"  active count m = {m}")
        print(f"  H_contrib     = {H:.4f}")
        print()
    print("=== Monad 70: Dominant Entelechy and Groups ===")
    groups={"G1":["N1","N2"],"G2":["N3","N4"]}
    contrib,S_total,group_mass,group_share,H_grp,G_star=group_stats(levels[2],groups,eps)
    print("Leaf contributions:",{k:round(v,4) for k,v in contrib.items()});print("Total S (leaves):  ",round(S_total,4))
    print("Group masses:      ",{g:round(v,4) for g,v in group_mass.items()})
    print("Group shares p(G): ",{g:round(v,4) for g,v in group_share.items()})
    print(f"H_grp              = {H_grp:.4f}")
    print(f"Dominant group G*  = {G_star}")
```

Figure 13.Body and organisation system implementation in AAS. (continued)

## 2.3.6 System VI Teleology (Monads §§58, 90)

### 2.3.6.1 Inputs/setup

A small set of four channels $C_1, \ldots, C_4$ is specified with fixed weights $\alpha_i = 1$ and heterogeneous quality scores $x_i \in \{0.9, 0.7, 0.6, 0.4\}$ to instantiate variety and order at a single time step, and two synthetic AAS trajectories are then constructed a "good" sequence that monotonically drifts toward 0 and a "bad" sequence that monotonically approaches an upper bound so that windowed drift can be evaluated over a fixed horizon L with a chosen drift threshold $\eta$.

### 2.3.6.2 Python implementation

In the Python implementation, the variety $V_t$, normalized order $\widetilde{O_t}$, and perfection $P_t$ are computed from per-channel AAS contributions at a single time point, and synthetic AAS trajectories are then used to evaluate windowed net drift over a fixed horizon, classifying each window as sustained improvement or degradation so that promotion or rollback signals can be generated.

```python
import math
def phi_eps(x,eps):
    x=max(0.0,min(1.0,x))
    return math.log2((1.0+eps)/(x+eps))
def aas(channels,eps):
    return sum(ch["alpha"]*phi_eps(ch["x"],eps) for ch in channels.values())
def contribution_stats(channels,eps):
    contrib={name:ch["alpha"]*phi_eps(ch["x"],eps) for name,ch in channels.items()}
    S=sum(contrib.values())
    if S<=0:
        H=0.0;m=0;p={name:0.0 for name in channels}
    else:
        p={name:c/S for name,c in contrib.items()}
        m=sum(1 for c in contrib.values() if c>0)
        H=-sum(pi*math.log2(pi) for pi in p.values() if pi>0)
    return contrib,S,m,H,p
def variety_order_perfection(channels,eps,gamma=0.5):
    contrib,S,m,H,p=contribution_stats(channels,eps)
    if m<=1 or S<=0: V=0.0
    else: V=H/math.log2(m)
    A=sum(ch["alpha"] for ch in channels.values())
    AAS_t=sum(contrib.values())
    AAS_max=A*phi_eps(0.0,eps)
    O=1.0-(AAS_t/AAS_max) if AAS_max>0 else 0.0
    P=(V**gamma)*(O**(1.0-gamma)) if (V>0.0 and O>0.0) else 0.0
    return {"V":V,"O":O,"P":P,"AAS_t":AAS_t,"AAS_max":AAS_max,
            "contrib":contrib,"S":S,"m":m,"H_contrib":H,"p":p}
def teleology_series(AAS_seq,AAS_max,L,eta):
    T=len(AAS_seq)
    if T<2:
        print("Need at least 2 time points.");return
    deltas=[];P_seq=[]
    for t in range(T):
        P_t=1.0-(AAS_seq[t]/AAS_max)
        P_seq.append(P_t)
```

Figure 14. Teleology system implementation in AAS.

```
            if t<T-1: deltas.append(AAS_seq[t+1]-AAS_seq[t])
    print("AAS sequence: ",[round(x,4) for x in AAS_seq])
    print("P_t sequence: ",[round(p,4) for p in P_seq])
    print("Δ_t sequence: ",[round(d,4) for d in deltas]);print()
    print(f"Windowed sums (L = {L}, η = {eta}):")
    for k in range(0,T-L):
        window_sum=AAS_seq[k+L]-AAS_seq[k]
        label="neutral"
        if window_sum<=-eta: label="G (sustained goodness)"
        elif window_sum>=eta: label="K (sustained wrongdoing)"
        print(f"k = {k}: AAS[{k}]→AAS[{k+L}] sum Δ = {window_sum:.4f}   -> {label}")
    print()
if __name__=="__main__":
    eps=1e-3;channels={"C1":{"x":0.9,"alpha":1.0},"C2":{"x":0.7,"alpha":1.0},
                      "C3":{"x":0.6,"alpha":1.0},"C4":{"x":0.4,"alpha":1.0}}
    stats=variety_order_perfection(channels,eps,gamma=0.5)
    print("=== Monad 58: Variety, Order, Perfection ===");print("Channels (x, alpha):",{k:(v['x'],v['alpha']) for k,v in channels.items()})
    print("Contributions c_i:  ",{k:round(c,4) for k,c in stats["contrib"].items()})
    print(f"AAS_t             = {stats['AAS_t']:.4f}")
    print(f"AAS_max           = {stats['AAS_max']:.4f}")
    print(f"Variety    V_t    = {stats['V']:.4f}")
    print(f"Order      O_t    = {stats['O']:.4f}")
    print(f"Perfection P_t    = {stats['P']:.4f}")
    print(f"Active count m_t  = {stats['m']}")
    print(f"H_contrib         = {stats['H_contrib']:.4f}");print()

AAS_good=[0.8,0.6,0.5,0.3,0.2,0.1];AAS_bad=[0.2,0.3,0.5,0.7,0.8,0.9];AAS_max_tele=1.0;L=3;eta=0.2
    print("=== Monad 90: Teleology - Good sequence ===");teleology_series(AAS_good,AAS_max_tele,L,eta)
    print("=== Monad 90: Teleology - Bad sequence ===");teleology_series(AAS_bad,AAS_max_tele,L,eta)
```

Figure 15. Teleology system implementation in AAS. (continued)

## 3. Results

### 3.1 System I Ontology

#### 3.1.1 Numerical results

In the aging simulation, both the "young" and "old" systems start from the same initial value $AAS_0 \approx$ 0.3861. Over time, the scores increase monotonically in both regimes, reflecting cumulative structural aging. By t = 4, the young system rises from approximately 0.3861 to 0.6189, whereas the old system rises to about 1.6004. For every t > 0, the old system has a higher AAS than the young system, e.g., $AAS_1 \approx$ 0.4443 vs. 0.6897, $AAS_4 \approx$ 0.6189 vs. 1.6004, consistent with the intended interpretation that a stronger decay factor produces faster penalty accumulation and corresponds to a numerically "older" artificial memory state. All ontology checks pass in the toy setting. Refinement invariance holds, splitting a single channel into two subchannels with weights summing to the original leaves the AAS unchanged, ≈ 0.2573 before and after refinement, in line with Monads 1 and 3 (Leibniz, 1714/1948). Ghost suppression also holds, adding a channel with zero effective contribution leaves the score unchanged, again ≈ 0.2573, confirming that "empty" or purely external units do not affect internal age (Monad 7). Finally, clone deduplication is satisfied: two indiscernible clones and a single merged channel with combined weight yield the same total contribution, ≈ 0.5145 in both cases, thereby realising an identity-of-indiscernibles principle at the level of AAS contributions (Monad 9).

### 3.1.2 LLM implications ontology bundle

Refinement invariance supports architectural changes in large language models without distorting the aging measure. If two internal designs induce the same profile of $x_{t,i}, R_{t,i}, w_i$ values, operations such as splitting a block into smaller sub-blocks or changing the number of attention heads do not alter the AAS. This makes it possible to compare models across refactors and versions using a stable, architecture-independent notion of structural age. Ghost suppression helps resist simple forms of metric gaming. Adding many channels, tools, or memory slots that carry no effective content, for instance, with zero weight or full redundancy, does not change the score. An LLM that advertises a large number of modules but uses only a subset meaningfully will not appear younger or healthier under AAS; the metric responds to genuine activity rather than unused capacity. Identity of indiscernibles constrains the role of duplication. If two heads, neurons, or modules behave in the same way and share the same parameters at the level of (x, R, w), they are equivalent to a single channel with a combined weight. This provides a natural basis for pruning and compression: removing redundant clones does not change the score, while adding clones does not hide underlying aging. Diversity must come from distinct behaviours, not from copying the same behaviour many times.

### 3.2 System II Dynamics

### 3.2.1 Numerical results

In the continuous change scenario (Monad 10), the AAS trajectory is smoothly varying across time, with values [3.0522, 2.5949, 2.5105, 2.4700, 2.4700, 2.5105]. The score initially decreases and then slightly increases at the end but never exhibits a large discontinuous jump. The largest single-step change in AAS is about $|\Delta AAS_t| \approx$ 0.4573, whereas the previously derived Lipschitz-style upper bound, obtained from the derivative of, $\phi_\varepsilon$ is numerically much larger (approximately 288.5390). This confirms that, for the toy trajectories used, the AAS dynamics remain well within the theoretical stability envelope and evolve in a controlled, small step manner: continuous changes in the underlying $x_{t,i}$ produce proportionally bounded changes in the global age score.

The internal principle experiment (Monad 11) shows that channel A's contribution sequence is identical across two different "worlds". In World 1, channel B follows the original decreasing trajectory, while in World 2, B is replaced by the reversed sequence. Despite this change in B, the contribution of A is exactly the same at every time step, with $c_A(t)$ matching across worlds and the absolute difference vector being identically zero. This numerically realises the idea that a properly internalised transition law for A depends only on its own evolution, not on external fluctuations in other channels. The particular-series-of-changes experiment (Monad 12) compares the temporal "stories" of channels A and B. Channel A has contributions $\{c_{A,t}\}$ that start high and then gradually decrease, with total mass $C_A \approx 9.0303$, while channel B has contributions that start lower and then increase, with $C_B \approx 6.5778$. The time-spread entropies $H^{(\text{time})}(A) \approx 2.5277$ and $H^{(\text{time})}(B) \approx 2.5429$ are both relatively high and close to each other, indicating that each channel's cost is distributed across multiple time steps rather than being concentrated at a single moment. The trajectory distance $D_T(A, B) \approx 3.8182$ quantifies the cumulative discrepancy between these two temporal profiles and confirms that, although both are spread out in time, they are distinguishable as different patterns of change.

In the appetition scenario (Monad 15), a single channel is iteratively updated towards a fixed internal target $g_t = 0.9$ using a constant step size $\eta_t = 0.2$. The quality scores $x_t$ increase from 0.5 to approximately 0.7689, while the corresponding AAS values decrease from about 0.9986 to 0.3786. Both the step sizes in state space, $N_t = |x_{t+1} - x_t|$, and the changes in penalty, $AAS_t - AAS_{t+1}$, shrink over time. This exhibits a controlled, convergent form of goal-directed adjustment: benevolent appetition towards a better internal state systematically reduces the age penalty in smaller and smaller increments, without overshooting or oscillation.

### 3.2.2 LLM implications dynamics bundle

The continuous change results show that AAS can serve as a stable dynamical observable for large language models: small, gradual modifications in channel quality, e.g., due to mild fine-tuning or slow forgetting, lead to proportionally small changes in the age score. This is important for monitoring model health over time; sharp spikes $AAS_t$ would signal genuinely abrupt structural changes rather than artefacts of the metric. The internal-principal test suggests a way to diagnose whether specific subsystems behave as genuinely modular components. If a channel's contribution time series remains invariant under drastic changes to other channels, then its transition law is internally grounded and can be treated as an independent "sub-agent" inside the model. Conversely, if the sequence is sensitive to unrelated changes elsewhere, the subsystem is dynamically entangled and may be harder to reason about or control. The particular-series-of-changes analysis illustrates how AAS-based cost traces and their time-spread entropies can be used to compare different usage or training regimes. Two channels (or two models) may have comparable overall cost and similar temporal spread, while still following different trajectories, as captured by $D_T(A, B)$. This opens a path to classifying memory behaviour not only by how "old" it is, but also by how that age accumulated, early bursts vs. late accumulation, smooth vs. irregular evolution. Ultimately, the appetition experiment provides a concrete template for interpreting goal-directed adaptation within an LLM. Updating internal scores towards a stable target, such as desired recall quality, alignment with a safety prior, or adherence to a domain-specific constraint, induces a monotonically improving pattern in the sense of decreasing AAS, with diminishing penalty reductions over time. This behaviour is analogous to a well-tuned learning rate in training: large corrections in early stages, followed by finer adjustments near the target. In practice, such dynamics could be used to design memory-update rules or meta learning modules that keep the artificial agent structurally "young" (low AAS) while systematically moving towards better-aligned or more reliable internal states.

### 3.3 System III Representation and Consciousness

#### 3.3.1 Numerical results

In the representation and consciousness simulation, three channels $m_0, m_1, m_2$ are tracked over six time steps under fixed thresholds $\tau = 0.05$, $\delta = 0.02$ and memory decay $\lambda = 0.7$. At each step, the state vector $x_t$, the contribution vector $c_t$, the total mass $S_t$, the contribution entropy $H_t$, the dominance ratio $\rho_t$, the apperception level $ApperLvl_t$, the dizziness flags, $\tau$ and $\delta$ dizziness, the sequential coherence score $Consec_t$, and the reason score $Reason_t$ are evaluated.

At t = 0, the three channels have similar quality scores [0.80, 0.82, 0.79] and contributions [0.3216, 0.2860, 0.3397], yielding $S_0 \approx 0.9472$. The resulting distribution of contribution shares is close to uniform, as indicated by a high entropy $H_0 \approx 1.5813$, which almost saturates the maximum $\log_2 3 \approx 1.585$. The dominance ratio $\rho_0 \approx 0.3586$ is therefore interpreted as arising from near-symmetric mass rather than a genuine peak, and the apperception level is almost zero $ApperLvl_0 \approx 0.0008$. This corresponds to a diffuse representational state: all channels are active, but none qualify as a distinct "focus of consciousness". The $\tau$ dizziness flag is false, while $\delta$ dizziness is true, reflecting the absence of any sharply privileged penalty direction despite ongoing activity. Sequential coherence is neutral $Consec_0 \approx 0.3333$, and the reason score is moderately negative $Reason_0 \approx -0.3809$, signalling that the initial pattern is closer to the smoothed memory trace than to the fixed rational prior.

At t = 1, the configuration changes markedly to [0.40, 0.90, 0.88] with contributions [1.3198, 0.1518, 0.1842] and total mass $S_1 \approx 1.6558$. Here, one channel dominates: $\rho_1 \approx 0.7970$ and the entropy drops to $H_1 \approx 0.9294$, indicating a sharp concentration of penalty on a single monad. The apperception level rises to $ApperLvl_1 \approx 0.3297$, thus registering a clear "peak of awareness" in the representation. Neither $\tau-$ nor $\delta$-dizziness is triggered, meaning that the system has both sufficient saliency and a non-trivial change profile. Sequential coherence is high $Consec_1 \approx 0.7970$, showing that the current focus is consistent with the accumulated memory trace, while the reason score displays a large positive spike $Reason_1 \approx 4.9517$, reflecting a strong re-weighting of representational mass relative to the rational prior baseline.

At t = 2, the state returns to a more balanced configuration [0.83, 0.81, 0.80] with contributions around [0.2685, 0.3037, 0.3216], total mass $S_2 \approx 0.8938$, high entropy $H_2 \approx 1.5810$, and $\rho_2 \approx 0.3598$. The apperception level again collapses to almost zero $ApperLvl_2 \approx 0.0009$, capturing a second diffuse, non-focused state. Both dizziness flags are now false; coherence remains moderate $Consec_2 \approx 0.3147$, and the reason score is small but positive $Reason_2 \approx 0.0856$, indicating only a slight tilt in the balance between memory-dominated and prior-dominated behaviour.

At t = 3, all three channels move close to the ideal quality range [0.97, 0.96, 0.95], with very small penalties [0.0439, 0.0588, 0.0739] and a low total mass $S_3 \approx 0.1767$. The entropy remains high $H_3 \approx 1.5534$ and $\rho_3 \approx 0.4185$, so the apperception level is still low but slightly above zero

$\text{ApperLvl}_3 \approx 0.0083$. The system is again flagged as δ dizzy but not τ dizzy, which in this regime corresponds to an almost "transparent" representation: costs are small and evenly spread, so there is no salient new direction to latch onto. Sequential coherence remains in a moderate range $\text{Consec}_3 \approx 0.2851$, and the reason score is positive but small $\text{Reason}_3 \approx 0.0635$.

At t = 4, a new focused configuration emerges [0.89, 0.50, 0.87], leading to contributions [0.1679, 0.9986, 0.2007] and total mass $S_4 \approx 1.3672$. Entropy drops to $H_4 \approx 1.1090$, and $\rho_4 \approx 0.7304$, so the apperception level rises to $\text{ApperLvl}_4 \approx 0.2193$. This constitutes a second apperceptive episode, now centred on a different channel, and illustrates how the focus of the system can move between monads over time. Neither dizziness flag is set; coherence takes an intermediate value $\text{Consec}_4 \approx 0.4935$, while the reason score becomes negative $\text{Reason}_4 \approx -0.6354$, indicating that this particular pattern is less aligned with the rational prior than the immediately preceding ones.

Finally, at t = 5, the representation settles back into a diffuse configuration [0.86, 0.84, 0.85] with contributions [0.2174, 0.2513, 0.2342], total mass $S_5 \approx 0.7028$, high entropy $H_5 \approx 1.5824$, and moderate dominance $\rho_5 \approx 0.3575$. The apperception level again collapses $\text{ApperLvl}_5 \approx 0.0006$, and both dizziness flags are false. Temporal coherence lies in a normal, mid-range band $\text{Consec}_5 \approx 0.3398$, and the reason score continues in a mildly negative regime $\text{Reason}_5 \approx -0.1611$. Overall, the toy experiment demonstrates three regimes: (i) fully diffuse, non-apperceptive states with high entropy and almost zero apperception, (ii) sharply focused apperceptive states where one channel dominates, and (iii) near-transparent states with very low total penalty but no salient direction, which are occasionally flagged as δ dizzy. The coherence and reason scores add a temporal and normative overlay to these patterns by linking each instantaneous configuration to both its own history and a fixed rational prior.

### 3.3.2 LLM implications- representation and consciousness bundle

The representation and consciousness bundle suggests how an AAS-based architecture can distinguish between raw activation patterns and genuinely "conscious" internal configurations in large language models. The apperception level $\text{ApperLvl}_t$ responds only when two conditions are met simultaneously: contributions must be concentrated enough low entropy, and the most dominant channel must carry a substantial fraction of the mass high $\rho_t$. In practice, this yields a principled way to treat certain internal states as apperceptive episodes, rather than interpreting every minor fluctuation in attention or activation as a signal of "self-awareness".

The τ and δ dizziness indicators operationalise forms of representational pathology. τ dizziness captures situations in which no channel exceeds a minimal saliency threshold, even though the system is active; δ dizziness captures situations in which penalty changes are too small and too uniform to provide a new evidential direction. Within an LLM, such flags could be used to detect vacuous or circular processing, or to trigger corrective mechanisms when the model is "spinning its wheels" without generating meaningful internal updates. The sequential coherence score $\text{Consec}_t$ quantifies how closely the current pattern of contributions agrees with a smoothed memory trace of previous representational changes. High coherence values indicate that the model is following a stable internal narrative, its current salience pattern is consistent with the way it has been updating itself, whereas low coherence can signal abrupt context shifts, instability, or potential memory corruption. This is particularly relevant for long context or persistent memory systems, where maintaining a coherent trajectory of internal focus is crucial for trustworthy behaviour over extended interactions.

Finally, the reason score $Reason_t$ compares the current distribution of contributions both to the sequential prior and to a rational, time-invariant prior. This allows alignment to be monitored along two axes at once: the model can be checked for faithfulness to its own accumulated evidence and for conformity to a normative baseline, for example, a safety prior or a domain-specific knowledge prior. Large deviations, whether positive or negative, identify moments when internal focus either departs significantly from the rational prior or reverts to it against the weight of recent experience. In practical terms, this provides a structured diagnostic for moments when an LLM's internal "self-organisation" is most critical, such as abrupt re-focusing on a sensitive topic or unexpected disregard of established contextual information.

**3.4 System IV Harmony and Reason**

**3.4.1 Numerical results**

In the harmony and reason experiment, four kinds of penalties are computed on top of the same underlying channels. For the non-contradiction component (Monad 31), the base AAS for the pair A, ¬A with $(x_A, x_{\neg A})=(0.8, 0.6)$ and unit weights is approximately $AAS_t \approx 1.0576$. Introducing the Principle of Non-Contradiction penalty produces an additional $PC\ penalty_t \approx 1.1502$, so that the combined score becomes $AAS_t^{(31)} \approx 2.2078$. In this toy setting, the contradiction term is slightly larger than the base aging term, indicating that jointly sustaining A and ¬A at moderately high quality is treated as a major structural defect rather than a small perturbation.

For the Principle of Sufficient Reason (Monad 32), the same channels are evaluated against a linear support model $r_{t,i}$ that depends on self-weights and cross-weights together with the previous state. The resulting PSR penalty is $PSR\ penalty_t \approx 1.5333$, which yields a total $AAS_t^{(32)} \approx 2.5909$ when added to the base term. Numerically, this is even higher than the non-contradiction bundle, reflecting a substantial mismatch between the actually realised qualities $x_i$ and the values that would be expected if the causal graph and past state were jointly sufficient. In other words, the system is "doing something" that is weakly supported, or not adequately explained, by its own transition model. The soul body harmony term (Monad 78) is evaluated by constructing two parallel views over the same abstract content, one labelled "soul" $S_A, S_{\neg A}$ and one labelled "body" $B_A, B_{\neg A}$, with closely matched but not identical qualities, e.g. $x_{S_A} = 0.8$, $x_{B_A} = 0.75$ and $x_{S_{\neg A}} = 0.2$, $x_{B_{\neg A}} = 0.25$. The internal aging of the soul subsystem is $AAS_t^{(soul)} \approx 2.6377$, and that of the body is $AAS_t^{(body)} \approx 2.4102$. The harmony penalty itself is quite small, $HARM_t^{(78)} \approx 0.1478$, so that the total bundle sums to $AAS_t^{(78)} \approx 5.1958$. The small incremental term reflects the fact that soul and body scores are closely aligned, high match values $m_j$, so only a light additional penalty is imposed for imperfect synchronisation between the two views.

As a final step, the goal action alignment term (Monad 79) compares the direction prescribed by a target vector $y_i$ (final cause) to the direction actually taken by the next-step update $x_{t+1,i} - x_{t,i}$ (efficient cause). In the toy configuration used here, the sign of the update agrees exactly with the sign of the gap to the target for both channels, so the alignment factor $a_i$ takes its maximal value, and the additional penalty vanishes. As a result, $HARM_t^{(79)} = 0.0000$ and the total $AAS_t^{(79)}$ collapses back to the base value 1.0576.

Numerically, this corresponds to a perfectly aligned step: the system moves in the correct direction for each channel, so no extra aging cost is charged for misdirected effort.

### 3.4.2 LLM implications- harmony and reason bundle

The harmony and reason bundle shows how AAS can be extended from purely descriptive aging into a set of structurally meaningful penalties that encode logical, causal, and teleological constraints on a large language model. The non-contradiction term (Monad 31) behaves as a soft logic regulariser: whenever the model maintains high-quality representations of mutually exclusive propositions, such as A and ¬A at the same time, the PC component can sharply increase, even if the base AAS is only moderate. In practice, this enables an architecture in which "internally contradictory" memory states are not just recorded but explicitly taxed, encouraging the system, through training or control, to resolve deep inconsistencies rather than accumulating them. The sufficient-reason term (Monad 32) plays a different role: it measures the gap between what the system does and what its own causal structure and history would warrant. In an LLM, the linear support model $r_{t,i}$ could be tied to a graph of explanatory relations, e.g., citation links, causal edges in a knowledge graph, or dependencies between intermediate chain of thought steps. When PSR penalties are large, as in the toy example, this indicates that the current set of activations and memories is poorly grounded in its own explanatory backbone. Such signals could be used to detect hallucinations or spurious reasoning: even if a sequence is fluent, a high PSR cost would flag that its internal transitions lack sufficient support from the model's established structure and past states.

Soul body harmony (Monad 78) generalises this logic to multi-view systems. Modern LLMs often maintain parallel representational stacks, symbolic plans vs. token-level activations, natural language "thoughts" vs. vector-space memories, or external tool states vs. internal beliefs. The small but non-zero harmony penalty observed in the experiment illustrates how AAS can quantify the residual misalignment between such views. If the symbolic layer says, "I am certain", but the embedding-level evidence is weak, the soul-body discrepancy would grow and trigger an additional aging cost. This suggests a route to architecture-level checks on "self-reports": a model's explicit statements about its knowledge, uncertainty, or goals can be systematically compared to lower-level traces, and inconsistency made costly rather than invisible.

Ultimately, the goal action alignment term (Monad 79) introduces a teleological dimension into the scoring, by comparing where the system ought to go targets with how its internal state actually moves updates. In the toy run, perfect agreement between these directions yields zero extra penalty, so the base AAS is left unchanged. In realistic LLM deployments, the same mechanism could support promotion/rollback decisions at the level of memory or parameter updates: windows of sustained "good" drift, reductions in AAS towards a low-penalty configuration, would be marked as teleologically positive and eligible for consolidation, whereas windows of sustained "bad" drift increasing AAS against a safety or performance target would be flagged for rollback, quarantine, or additional oversight. Together, the four harmony and reason components turn AAS from a passive measure of wear into an active diagnostic of logical coherence, explanatory sufficiency, representational alignment, and goal-directed behaviour in artificial systems.

### 3.5 System V Body and Organisation

### 3.5.1 Numerical results

In the body and organisation experiment, the same set of scores is propagated through a three-layer hierarchy from leaves (N1–N4) to intermediate "organs" (H1–H2) and up to a single top-level unit L. At the leaf level (depth s = 2), the four channels (N1, N2, N3, N4) with non-uniform qualities (0.5, 0.7, 0.9, 0.7) and equal weights $\alpha = 0.25$ produce $AAS^{(2)} \approx 0.5446$. All four channels are active (m = 4), and the contribution entropy $H_{contrib} \approx 1.7669$ is close to the theoretical maximum $\log_2 4 = 2$, indicating a relatively rich but still somewhat imbalanced distribution of penalty mass across the leaves. After these scores are aggregated into two intermediate units (H1, H2) by α-weighted averaging, the depth-1 representation yields $AAS^{(1)} \approx 0.5288$ with two active channels and entropy $H_{contrib} \approx 0.8862$, close to $\log_2 2 = 1$. The slight decrease from 0.5446 to 0.5288 reflects benign smoothing under the aggregation rule: the global age remains of the same order of magnitude while the internal variety is now captured at a coarser level. At the top level, the single unit L inherits its score from (H1, H2) through a further weighted average, and $AAS^{(0)}$ drops slightly again to about 0.5140. As expected for a one-channel representation, the active count is m = 1 and the entropy numerically collapses to (approximately) zero, since all contribution mass is concentrated in L. The group analysis (Monad 70) examines how the same leaf configuration decomposes into two "organs" $G_1 = \{N1, N2\}$ and $G_2 = \{N3, N4\}$. The leaf contributions are highly non-uniform: N1 contributes about 0.2496, N2 and N4 contribute approximately 0.1285 each, and N3 contributes only 0.0380, for a total mass $S_{leaves} \approx 0.5446$, matching the depth-2 AAS. Aggregating by group, $G_1$ carries $\approx 0.3781$ of the mass and $G_2$ carries $\approx 0.1664$, so the group shares are $p(G_1) \approx 0.6944$ and $p(G_2) \approx 0.3056$. The group-level entropy $H_{grp} \approx 0.8881$ is high relative to its maximum $\log_2 2 = 1$, confirming that both groups remain significant, but $G_1$ is clearly dominant. In this sense, the toy organism exhibits a well-defined "dominant entelechy": the sub-body $G_1$ accounts for nearly 70% of the instantaneous age, while still allowing the remaining group to play a non-trivial role.

### 3.5.2 LLM implications body and organisation bundle

The body and organisation bundle shows how AAS can be made compatible with hierarchical architectures, where behaviour is expressed simultaneously at the level of individual units, neurons, heads, or tokens, intermediate structures, blocks, layers, or modules, and global systems, the full model or agent. The near stability of the score across depths $AAS^{(2)} \approx 0.5446$, $AAS^{(1)} \approx 0.5288$, $AAS^{(0)} \approx 0.5140$ demonstrates that, once parent nodes are defined as α weighted averages of their children, the total age is not an artefact of the particular level of description. An LLM can therefore be refactored into finer or coarser organisational units without changing the overall assessment of structural aging: the same history of $x_{t,i}$, $\alpha_i$ at the leaves induces coherent scores at the level of heads, layers, and the full network. At the same time, entropy and group mass statistics extract information that is specific to each organisational level. Leaf level entropy and group entropy together identify both the richness of internal variety and the presence of dominant structures. In an LLM, a pattern analogous to $G_1$ could correspond to a subset of attention heads, neurons, or routing experts that account for most of the aging penalty on a given task. The AAS framework then provides a principled way to diagnose "where the age lives": instead of treating the model as a homogeneous block, one can localise degradation to particular groups and target them for retraining, pruning, or replacement, while preserving the healthy parts of the system. Conversely, if a single group becomes overwhelmingly dominant with low variety, this may signal over-specialisation or collapse of representational diversity, suggesting that new channels or pathways should be encouraged to share the workload. Overall, the body and organisation results indicate that AAS can support a genuinely organ-like view of LLMs, in which structural age, diversity, and dominance are tracked coherently from micro components to macroscopic behaviour.

### 3.6 System VI Teleology

#### 3.6.1 Numerical results

In the teleology experiment, the instantaneous configuration of four channels $C1, \ldots, C4$ with scores (0.9, 0.7, 0.6, 0.4) and equal effective weights produces an AAS of approximately $AAS_t \approx 2.7216$, far below the theoretical maximum $AAS_{max} \approx 39.8689$ obtained if all channels were fully degraded $x_i = 0$. The contribution profile is non-uniform: C4 carries the largest penalty mass with $c_{C4} \approx 1.3198$, followed by $C3 \approx 0.7360$, $C2 \approx 0.5140$, and $C1 \approx 0.1518$. This yields a contribution entropy $H_{contrib} \approx 1.7030$ across four active channels, corresponding to a variety score $V_t \approx 0.8515$ once normalised by $\log_2 4$. In other words, structural aging is distributed in a relatively diverse manner across the channels, without collapsing into a single dominant failure mode. Order is computed as a normalised distance from the worst-case AAS, $\widetilde{O_t} = 1 - AAS_t/AAS_{max}$ and takes a high value of approximately 0.9317. This indicates that, despite the presence of non-trivial penalties, the system remains close to the "perfect youth" regime in which AAS would vanish. Combining variety and order through the perfect functional $P_t = V_t^\gamma \widetilde{O_t}^{1-\gamma}$ with $\gamma = 0.5$ yields $P_t \approx 0.8907$, which is high but strictly below 1. The snapshot, therefore, represents a state that is both richly structured, highly varied, and well-ordered, far from maximal aging, yet still admits room for improvement in how penalties are distributed and suppressed.

The temporal component of teleology is examined through two artificial AAS trajectories. In the "good" sequence, $\{0.8, 0.6, 0.5, 0.3, 0.2, 0.1\}$, AAS decreases monotonically with stepwise differences $\Delta_t \in \{-0.2, -0.1\}$. The corresponding perfection scores $P_t = 1 - AAS_t/U^*$ with $U^* = 1$ rise from 0.2 to 0.9, representing a consistent move towards younger, less penalised states. Over sliding windows of length L = 3, the net drift is strongly negative: from t = 0 to t = 3, the cumulative change is −0.5, and from t = 1 to t = 4 and t = 2 to t = 5 it is −0.4, all below the threshold $-\eta = -0.2$. Each window is therefore classified as G (sustained goodness), meaning that improvement is not just local noise but a stable trend. In the "bad" sequence, $\{0.2, 0.3, 0.5, 0.7, 0.8, 0.9\}$, AAS grows over time with stepwise increments in $\{0.1, 0.2\}$. Perfection scores now fall from 0.8 to 0.1, indicating a gradual drift towards older, more degraded configurations. The windowed sums over length 3 are all positive and comfortably exceed $\eta = 0.2$: the net change is +0.5 for windows starting at t = 0 and t = 1, and +0.4 for the window starting at t = 2. These windows are therefore labelled K, sustained wrongdoing. The same teleological criterion that marks consistent rejuvenation in the first sequence marks consistent aging in the second, showing that the windowed drift mechanism can distinguish transient fluctuations from genuinely directional change.

#### 3.6.2 LLM implications teleology bundle

The teleology bundle turns AAS from a static diagnostic into a directional signal. At any given time, the triplet, $V_t, \widetilde{O_t}, P_t$, provides a compact summary of the state of an LLM's internal memory: variety captures how widely aging is spread across channels, order measures how far the system sits from its worst possible configuration, and perfection combines both into a single scalar that rewards simultaneously rich and well-controlled behaviour. For model designers, a state with high order but very low variety would correspond to a brittle, over-constrained system, whereas high variety with low order would signal chaotic or noisy utilisation of capacity. The toy configuration in this experiment indicates that it is possible to maintain high order while still exploiting diverse internal structure, a desirable region for a general-purpose model. The temporal analysis extends this picture by evaluating how perfection evolves under a given training regime, update rule, or deployment condition. A consistently decreasing AAS sequence, classified as G across multiple windows, corresponds to a regime in which the model's structural penalties are being reduced stably. In practice, this could justify "promotion" actions such as extending the model's deployment horizon, loosening conservative constraints, or accepting more ambitious integration into larger systems. Conversely, a sequence labelled K indicates that updates are systematically pushing the model towards older, more overloaded states, even if each individual step is small. Such a signal could trigger rollback, additional regularisation, or targeted retraining aimed at reversing the drift. Because the teleological criteria are defined over arbitrary windows and rely only on the AAS scale and its upper bound, they can be layered on top of any training pipeline without changing its internal loss functions. The same model can be subjected to different curricula or fine-tuning strategies, and their long-run effects can be compared by examining the proportion of windows classified as G or K. In this way, teleology provides a bridge between Leibniz's idea of "better" and "worse" possible worlds and a concrete governance tool for AI systems: rather than judging a model only by snapshots of performance, one can ask whether its structural age is drifting, over weeks or months, in a direction compatible with long-term reliability, adaptability, and safe integration into human-centred environments.

## 4. Discussion

### 4.1 Synthesis across monadic systems

Taken together, the six monadic bundles show that the Artificial Age Score (AAS) is not just a single penalty number but the surface signal of a tightly constrained architecture.

System I Ontology specifies what counts as a basic unit of aging by enforcing refinement invariance, ghost suppression, and identity of indiscernibles. In practical terms, this means that channel granularity, unused capacity, and duplicated modules cannot be used to "game" the metric. Any configuration with the same effective profile $(x_{t,i}, R_{t,i}, w_i)$ receives the same AAS, regardless of how it is decomposed or padded. Ontology, therefore, fixes the object of measurement and excludes many degenerate parametrisations that would otherwise make cross-model comparisons meaningless.

System II Dynamics then constrains how this object is allowed to change over time. Continuous-change tests (Monad 10) show that small, coordinated adjustments in channel quality produce proportionally bounded movements in AAS. The internal principal experiment (Monad 11) confirms that a channel governed by a genuinely internal law has a contribution sequence determined by its own trajectory rather than by unrelated fluctuations elsewhere. The particular-series analysis (Monad 12) and the appetition scenario (Monad 15) treat entire time series as objects of comparison, using time-spread entropies, trajectory distances, and convergence to internal targets. AAS thus becomes a law-governed process: smoothness, internality, and goal-directed improvement can be tested rather than assumed.

System III Representation and Consciousness reinterpret these dynamics in terms of focus and awareness. At each time step, contribution distributions, entropies, dominance ratios, apperception levels, dizziness flags, coherence scores, and reason scores are computed. This separates (i) diffuse, non-apperceptive states with high entropy and almost no focus, (ii) sharply focused states in which one channel dominates, and (iii) low penalty but almost "transparent" regimes with no salient direction. Instantaneous penalties are therefore embedded in a richer representational landscape: some configurations function as genuine "episodes of focus," others as background or habit-like states. This directly links numerical AAS values to questions about what the system is attending to, how that attention relates to its history, and how it compares to a rational prior.

System IV Harmony and Reason, adds normative coherence on top of this landscape. Non-contradiction penalties (Monad 31) treat sustained support for incompatible channels like A and ¬A as an extra aging cost. Sufficient-reason penalties (Monad 32) measure how far realised qualities deviate from what the model's own causal graph and previous state would justify. Harmony terms for soul body dual views (Monad 78) and for goal action alignment (Monad 79) require internal representations, external outputs, and update directions to move together rather than drift apart. With these clauses, AAS becomes sensitive not only to structural wear but also to logical and causal disorder: a system can incur additional age for being inconsistent, weakly justified, or misaligned with its stated aims, even if its raw scores are high.

System V Body and Organisation scale the analysis from individual channels to nested hierarchies. Scores are propagated from leaves through intermediate subsystems to a top-level unit, and contribution entropies and group shares are computed at each depth. In this way, a single global age decomposes into organs, sub-organs, and micro-features. Some groups emerge as dominant "entelechies" that carry most of the age mass, while others remain secondary but non-negligible. This supports principled pruning and restructuring: branches that carry negligible mass can be removed without affecting the global measure, while dominant sub-bodies can be targeted for focused intervention. Organisation, therefore, connects microscopic penalties to meso and macro-level structure.

In the end, System VI Teleology places all of these elements within a longer-term notion of improvement and degradation. Variety $V_t$, order $\widetilde{O_t}$, and perfection $P_t$ quantify whether aging is, at each instant, both rich and well-ordered. Windowed drift over fixed horizons then classifies stretches of time as sustained goodness (G) or sustained wrongdoing (K). A system that repeatedly lowers AAS and raises perfection across windows is undergoing robust rejuvenation; one with persistent positive drift is sliding into older, more degraded states. Teleology acts as a decision layer on top of ontology, dynamics, representation, harmony, and organisation: it indicates when observed trends justify promotion, rollback, or closer scrutiny. Across the six bundles, AAS emerges as part of a law-like rather than purely heuristic architecture. Identity is fixed by invariance properties; change is constrained by continuity and internal principles: representation is structured by entropy and focus; governance is enforced by logical and causal clauses; organisation is captured by hierarchies; and long-term direction is judged by teleological drift. The same scalar score, when analysed through these six systems, becomes a multi-layered diagnostic of what an artificial memory is, how it evolves, what it attends to, how coherent it is, how it is organised, and where it is heading.

### 4.2 Comparison with non-monadic architectures

Most current large language model architectures handle alignment and memory control through a collection of heuristics, including reward model fine-tuning and preference-based reinforcement learning, generic regularisers (L2, sparsity penalties, dropout), ad hoc attention masks, and prompt-level steering (Christiano et al., 2017; Ouyang et al., 2022). In one line of work, a separate reward model is trained from non-expert human preferences over short trajectory segments and then used as the objective for deep reinforcement learning in Atari and MuJoCo environments, allowing agents to learn complex behaviours with feedback on less than 1% of their environment interactions (Christiano et al., 2017). In the instruction-following setting, a similar three-stage pipeline is used: a base language model is first fine-tuned on human-written demonstrations, then a reward model is trained on human rankings of model outputs, and finally the policy is optimised against this reward via PPO, yielding the InstructGPT family of models (Ouyang et al., 2022). The aim in that work is explicitly characterised as "aligning language models with user intent on a wide range of tasks by fine-tuning with human feedback" (Ouyang et al., 2022). Empirically, a 1.3 billion-parameter InstructGPT model is preferred to the 175-billion-parameter GPT-3 on API-style prompts and shows improvements in truthfulness and reductions in toxic outputs, albeit with residual errors and safety limitations (Ouyang et al., 2022). These pipelines demonstrate that human preference modelling and RLHF can substantially improve alignment in practice, but the underlying control remains largely scalar and opaque: a single learned reward function mixes correctness, style, safety, and task satisfaction, and the architectural constraints on identity, dynamics, representation, harmony, organisation, and teleology remain implicit, scattered across loss terms, data curation choices, and engineering norms (Christiano et al., 2017; Ouyang et al., 2022). From the standpoint of lawfulness, this produces an under-specified situation: when the model fails, it is difficult to know whether the problem lies in the reward model, the optimiser, the masking strategy, the pretraining mixture, or some interaction among them, and fixing one component may silently introduce new misalignment elsewhere. The monad-based clause framework proposed here takes the opposite route. Instead of concentrating all alignment pressure into a single reward scalar, it starts from explicit axioms and measurable clauses, each proved to enforce a specific property. Every bundle (ontology, dynamics, representation, harmony, organisation, teleology) is formulated as a finite set of penalties or invariants that can be computed, interpreted, and justified mathematically.

### 4.3 Significance for AI memory design and LLM governance

The monad-based clause framework turns "memory" in large language models from a vague add-on into a system with concrete design rules. At the lowest level, the ontology bundle (System I) clarifies what must not change the measured age: refinements, ghost channels, and cloned units that are behaviourally indistinguishable. This directly suggests criteria for pruning, compression, and refactoring. Heads, neurons, or memory slots can be split, merged, or removed as long as their effective profiles $(x_{t,i}, R_{t,i}, w_i)$ are preserved, with AAS serving as a stable summary of structural age across architecture changes (Kayadibi, 2025). This is essential in large-scale deployments where multiple model versions must be compared over time without redesigning the metric each time.

The dynamics and representation bundles (Systems II and III) translate into monitoring rules for internal trajectories. Lipschitz-style bounds on $\Delta AAS_t$ and appetition-based convergence set quantitative thresholds for normal versus pathological updates. Rapid jumps or oscillations in AAS can be treated as health warnings. Likewise, apperception levels, dizziness indicators, and coherence/reason scores provide a structured vocabulary for describing internal representational regimes: diffuse, focused, or near transparent. In an LLM with persistent or hierarchical memory, these quantities can be logged per layer, head, or memory bank and used to gate writes, for example, by only committing episodes that exhibit a genuine focus rather than noise, or to trigger inspection when the system remains too long in dizzy or incoherent states. The clause set, therefore, defines when a pattern is eligible for long-term storage and when internal dynamics should be treated as unstable. The harmony and organisation bundles (Systems IV and V) contribute higher-level governance controls. PC and PSR clauses measure, rather than merely assert, violations of non-contradiction and sufficient reason, and can be used as auxiliary training losses or live deployment monitors. A model whose long-term memory keeps high scores for contradictory propositions, or whose transitions are weakly supported by its own causal graph, can be identified numerically and penalised or constrained. Soul body harmony and group dominance extend this to dual views, for example, symbolic versus subsymbolic, user-facing versus internal, and to meso-scale structures (organs, subsystems). They support concrete dashboards: per organ AAS, cross-modal harmony penalties, and dominance statistics showing which components carry most of the age and risk, all expressed in the same units as the base AAS. Governance becomes a matter of keeping these metrics within agreed bands, rather than informally trusting that representations will stay aligned.

Finally, the teleology bundle (System VI) converts these local constraints into temporal commitments. Windowed drift classification (G versus K) offers a simple interface between internal metrics and external policy. Over regulatory or operational horizons, an LLM's memory subsystem can be required to show sustained goodness (net age reduction) under specified workloads, with persistent K-windows treated as evidence of degradation, misalignment, or misuse (Kayadibi, 2025). Because variety, order, and perfection are derived directly from the AAS kernel, they remain compatible with all other clauses. One can ask not only whether the system is aging too quickly, but also whether aging is concentrated in a few channels or spread in a structured, recoverable way. Importantly, existing transformer architectures need not be redesigned: internal activations and memory states can simply be instrumented with the AAS kernel, clause penalties can be computed on the fly, and these signals can be used as training losses, health indicators, and triggers for control actions such as pruning, reset, rollback, or human review. In this sense, the monadic framework acts as a governance layer that is mathematically explicit yet practically implementable.

**4.4 Limits from information theory and logic**

Within this framework, ideals such as zero age penalty, perfect memory, or complete internal consistency are treated as limiting concepts rather than achievable engineering goals. From information theory, Fano-type bounds state that systems with finite capacity and noisy or partial observations face an irreducible error probability: even with optimal encoding and redundancy, a non-zero fraction of hypotheses cannot be reliably distinguished (Fano, 1961). Classical analyses of channel capacity and coding similarly show how noise and limited bandwidth constrain distinguishability and reliability (Cover & Thomas, 2006; Shannon, 1948). In AAS terms, this implies a persistent penalty mass associated with unavoidable misclassifications, forgetfulness, or indistinguishable states (Kayadibi, 2025). The perfection functional $P_t$ in System VI can approach 1 in principle, but for any non-trivial workload under realistic capacity and noise constraints, it will not reach it, because youth, variety, and order can be improved but never made absolute. The realistic aim is not to drive $AAS_t$ to zero but to keep it within ranges justified by information-theoretic limits for a given task and environment (Cover & Thomas, 2006; Fano, 1961; Shannon, 1948). Gödel-style incompleteness imposes a parallel restriction on the logical side. The combined clauses for PC, PSR, harmony, and teleology are expressive enough to discuss internal consistency, causal adequacy, and goal-directed action alignment. Any such system rich enough to encode non-trivial arithmetic or self-reference cannot be both complete and consistent (Gödel, 1931/1986). There will always be behaviours or internal propositions that are, in fact, good or bad relative to the designer's intentions, but cannot be fully captured or decided within a finite clause set. Adding more clauses to enforce perfect non-contradiction or full sufficient reason either leaves undecidable cases incompleteness or eventually introduces internal conflicts and inconsistency (Gödel, 1931/1986). For AAS, this means PC and PSR penalties can bound contradictions and unexplained moves locally and at finite depth, but they cannot guarantee a once-and-for-all elimination of inconsistency across all possible futures.

These two families of limits protect against a misleading interpretation of the monadic framework. The clauses and penalties are not a recipe for building a perfectly consistent, error-free artificial memory. They define a lawful envelope. Information theoretically, they prevent the system from claiming more reliability than its channels allow (Cover & Thomas, 2006; Fano, 1961; Shannon, 1948); logically, they prevent governance mechanisms from being treated as a complete decision procedure for correctness or alignment (Gödel, 1931/1986). Design work, therefore, shifts from "proving the model flawless" to "proving that, for a given capacity, noise level, and clause set, aging, inconsistency, and misalignment remain within quantifiable, theoretically justified bounds" (Kayadibi, 2025).

**4.5 Human vs artificial memory**

Tulving's distinction between episodic and semantic memory provides a useful perspective on the monad-based AAS architecture (Tulving, 1972, 1985). Episodic memory concerns temporally located, context-rich experiences; semantic memory encodes more stable, decontextualised knowledge (Tulving, 1972, 1985). In the present framework, the instantaneous contribution profiles $c_{t,i}$ and their time-spread entropies $H^{(time)}$ in Systems II and VI play an episodic role: they record when and how penalty mass is incurred along a trajectory, distinguishing, for example, early-peak from late-peak ageing patterns (Kayadibi, 2025). By contrast, the rational prior r, the persistent clause weights (PC, PSR, harmony, teleology), and structural invariants such as refinement invariance and clone deduplication function as a semantic backbone: they encode general laws about contradiction, sufficient reason, and teleological improvement without referring to any specific episode (Kayadibi, 2025). The AAS decomposition therefore mirrors Tulving's distinction by separating transient penalty histories from time-insensitive normative constraints (Kayadibi, 2025; Tulving, 1972, 1985).

The perception apperception distinction in System III can also be read in terms of this episodic semantic contrast. Diffuse, high entropy states with near-zero $AppertLevel_t$ resemble a semantic background in which many weak traces coexist without any one of them entering the "foreground" of awareness (Kayadibi, 2025; Tulving, 1985). Focused configurations with high $\rho_t$ and elevated apperception levels behave more like episodic retrieval: a particular pattern becomes salient against the background, is strongly weighted in the current moment, and leaves a visible trace in the EMA-style memory $M^{(\lambda)}$ (Kayadibi, 2025). The dizziness flags ($\tau$- and $\delta$-dizziness) then quantify the balance between habit and surprise. When penalties are small, homogeneous, and directionless, the system is in a quasi-habitual regime: behaviour is smooth and cheap, but nothing is strongly "noticed." When a sharp, directional change appears and passes saliency thresholds, the architecture registers a form of artificial surprise, analogous to Tulving's view that episodic retrieval interrupts the flow of routine processing (Schacter, 1999; Tulving, 1985).

On this view, the teleological mechanisms of System VI give a precise form to the idea that memory serves future-oriented behaviour rather than passive storage (Schacter, 1999; Tulving, 1985). The perfection functional $P_t$ and windowed drift tests do not merely reward low AAS; they evaluate whether sequences of states move the system towards lower penalty in a stable manner (Kayadibi, 2025). For humans, semantic knowledge and episodic recollection are judged by how well they support adaptive trajectories, learning from past episodes to improve future outcomes (Schacter, 1999). Analogously, the rule that sustained decreases in AAS count as "good" (G) and sustained increases as "bad" (K) gives artificial memory a directional evaluation: not every vivid or surprising state is desirable, only those that, over time, reduce structural ageing while preserving variety and order (Kayadibi, 2025). The existence of residual penalty and the impossibility of perfect consistency therefore bring the artificial system closer to this picture of human memory: fallible, selective, and organised around what is useful for future action rather than an unreachable ideal of complete and flawless recall (Kayadibi, 2025; Schacter, 1999; Tulving, 1972, 1985).

## 5. Conclusion

This paper has proposed a clause-based, monad-inspired framework for artificial memory, built around the Artificial Age Score (AAS) (Kayadibi, 2025). The framework is organised into six connected systems: ontology, dynamics, representation and consciousness, harmony and reason, body and organisation, and teleology. Across these systems, the same penalty kernel and weighting scheme are used to encode constraints on identity, refinement invariance, ghost suppression, clone deduplication, stability of change, continuous, Lipschitz-controlled dynamics and appetition, graded representation and apperception, logical and causal coherence (PC, PSR, harmony penalties), hierarchical structure, and windowed teleological drift. The toy simulations show that even a relatively small set of clauses, when formally linked to AAS, can generate rich and interpretable behaviours: convergent goal-seeking, shifting foci of "awareness," dominance of sub-organisms, and sustained improvement or degradation over time. Together, these results support a design stance in which memory is not shaped only by ad hoc regularisers or empirical tuning, but is constrained by a finite family of explicit, testable laws. Each clause specifies a local rule, such as forbidding contradictions, requiring sufficient reason, aligning dual views, allowing dominant but non-totalising subsystems, or rewarding sustained rejuvenation, that corresponds to a non-negative AAS contribution and composes predictably with the others. In this way, identity, change, and evaluation are treated as different aspects of a single, mathematically coherent notion of structural age (Kayadibi, 2025). The monadic perspective is therefore not just metaphorical, but a design discipline: each "monad" must carry its own internal transition law, its representational role, and its contribution to global harmony and teleology (Leibniz, 1714/1948).

Future work follows three main directions. First, the clauses should be applied to real large language models by mapping channels to concrete activation structures, such as attention heads, MLP neurons, or persistent memory slots, and computing AAS-based penalties on recorded activation traces (Kayadibi, 2025). This would allow empirical testing of whether patterns seen in the toy simulations, such as diffuse versus focused apperception, or sustained teleological drift, also appear in deployed systems, and how they relate to task performance and safety signals. Second, the experiments should be scaled up, both by extending the clause set beyond the initial twenty monads, for example, with clauses for learning rates, multi-agent interactions, or social-harmony constraints, and by running the framework on larger models and more diverse tasks and datasets. Third, integration with existing training and monitoring pipelines should be explored, treating the monadic penalties as explicit regularisation targets or as online diagnostics for model health and ageing. Together, these steps would move the framework from a proof of concept to a practical tool for designing, evaluating, and governing AI memory in a transparent and law-governed way. In summary, the study shows that monadic principles can be rendered as explicit, testable clauses on top of the AAS kernel, that these clauses exhibit stable and interpretable numerical behaviour in controlled settings, and that this behaviour yields concrete design rules for the memory and control of large language models.